\pdfoutput=1

\documentclass[11pt]{article}

\usepackage[preprint]{acl}

\usepackage{times}
\usepackage{latexsym}

\usepackage[T1]{fontenc}

\usepackage[utf8]{inputenc}

\usepackage{microtype}

\usepackage{inconsolata}
\usepackage{microtype}
\usepackage{hyperref}
\usepackage{url}
\usepackage{booktabs}
\usepackage{amsmath}
\usepackage{graphicx}
\usepackage{authblk}
\usepackage{stackengine}
\usepackage{adjustbox}
\usepackage{float}

\usepackage{graphicx}

\title{Using LLMs to Aid Annotation and Collection of Clinically-Enriched Data in Bipolar Disorder and Schizophrenia}

\author{
 \textbf{Ankit Aich\textsuperscript{1, 5, 6}},
 \textbf{Avery Quynh\textsuperscript{2}},
 \textbf{Pamela Osseyi\textsuperscript{5}},
 \textbf{Amy Pinkham\textsuperscript{3}},\\
 \textbf{Philip Harvey\textsuperscript{4}},
 \textbf{Brenda Curtis\textsuperscript{5}},
 \textbf{Colin Depp\textsuperscript{2}},
 \textbf{Natalie Parde \textsuperscript{1}},

\\
 \textsuperscript{1}Department of Computer Science, University of Illinois Chicago,\\
 \textsuperscript{2}University of California, San Diego, \\
 \textsuperscript{3}University of Texas Dallas,
 \textsuperscript{4}University of Miami,\\
 \textsuperscript{5} National Institute on Drug Abuse, National Institutes of Health,\\
 \textsuperscript{6}School of Engineering and Applied Science, University of Pennsylvania\\
\\
}

\begin{document}
\maketitle

\begin{abstract}

NLP in mental health has been primarily social media focused. Real world practitioners also have high case loads and often domain specific variables, of which modern LLMs lack context. We take a dataset made by recruiting 644 participants, including individuals diagnosed with Bipolar Disorder (BD), Schizophrenia (SZ), and Healthy Controls (HC). Participants undertook tasks derived from a standardized mental health instrument, and the resulting data were transcribed and annotated by experts across five clinical variables. This paper demonstrates the application of contemporary language models in sequence-to-sequence tasks to enhance mental health research. Specifically, we illustrate how these models can facilitate the deployment of mental health instruments, data collection, and data annotation with high accuracy and scalability. We show that small models are capable of annotation for domain-specific clinical variables, data collection for mental-health instruments, and perform better then commercial large models.
\end{abstract}

\section{Introduction}
\label{sec:intro}

The inherent complexity of mental health data presents significant challenges, even as the availability of AI systems designed to aid in its understanding and categorization continues to grow \citep{pam_paper_14}. AI-based systems have increasingly leveraged social media as a data source in the realm of mental healthcare, leading to the development of pre-trained models like MentalBERT \citep{ji2022mentalbert} and initiatives to classify and detect various mental health phenomena, such as schizophrenia \citep{liu2022linguistic}, disease progression \citep{psychotic_relapse_with_fb}, depression \citep{twitter_depression_1}, and stress \citep{twitter_stress_1}.

\begin{figure}[!ht]
    \centering
    \hspace{-1cm}
    \includegraphics[scale=0.35]{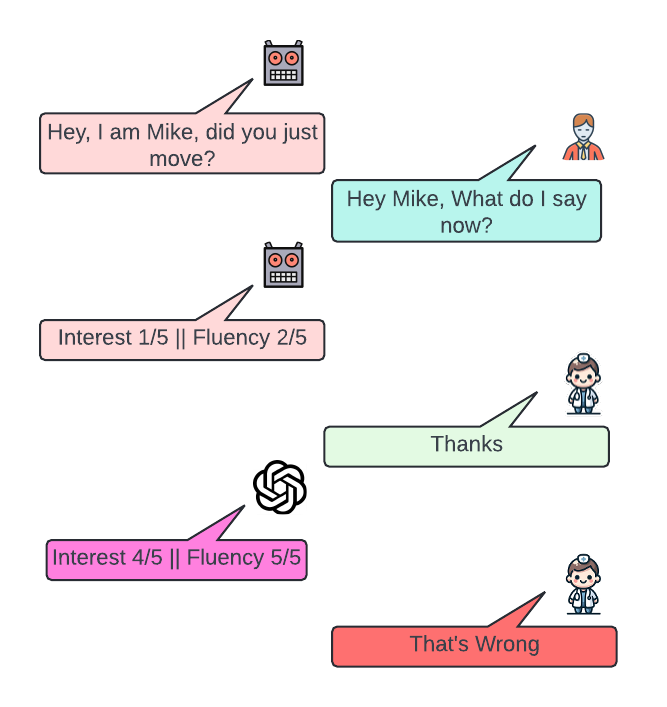}
    \caption{Our method creates a fine-tuned model. This model is able to directly interact with recruited participants to help them undertake established mental health instruments through turn-based tasks. It can annotate for clinical variables with low error. We see that commercial LLMs like GPT-4 / GPT-4o cannot annotate when it comes to clinical variables which are niche to a domain.}
    \label{fig:paper_diagram}
\end{figure}

In addition to the ethical issues surrounding the use of social media for clinical diagnoses, numerous other challenges persist. These include participant bias \citep{participant_bias}, issues with generalizability \citep{mitchell-etal-2015-quantifying}, and an overreliance on self-disclosure or non-clinical labels \citep{mitchell-etal-2015-quantifying, coppersmith-etal-2014-quantifying}.

Psychiatric disorders such as schizophrenia and bipolar disorder are often characterized by language deficiencies \citep{task_perform_schizo}. Individuals with these conditions may exhibit disorganized language comprehension and speech patterns \citep{language_schizo_disorga}. Consequently, text or speech-based mental health instruments can be employed to assess individuals with medically validated diagnoses, thereby elucidating the effects of psychiatric disorders.

Previous efforts to apply AI in the context of schizophrenia and bipolar disorder have predominantly focused on automated diagnoses using smaller datasets \citep{classify_schizo}. These classification endeavors have encountered multiple challenges. For instance, social media data often results in non-clinical labels \citep{twitter_schizo}, while the classification of clinical data is complicated by small datasets \citep{classify_schizo}, underutilization of records \citep{classify_schizo_2}, and attempts to apply AI to multiple psychiatric disorders simultaneously \citep{classify_OCD_schizo_BD}.

Moreover, the scarcity of robust data sources for mental health care and AI remains a significant barrier, as noted by \citet{harrigian-etal-2021-state}. The reliability of social media labels is further undermined over time due to evolving subjective annotation metrics \citep{harrigian-dredze-2022-now}. To enhance the application of AI and language models in schizophrenia and bipolar disorder research, we propose a novel approach. This approach involves testing the efficacy of AI models in the context of data collection and annotation.

Our study utilizes a dataset comprising 644 participants with established medical histories of schizoaffective disorder, schizophrenia (SZ), bipolar disorder (BD), or who are healthy controls. These participants undergo a mental health assessment involving interviews conducted by expert clinicians \citep{PATTERSON2001351}. We engaged two expert clinicians to annotate transcribed speech samples across five clinical variables. Importantly, we do not conduct automated diagnoses nor suggest that language models should be used for diagnostic purposes. Instead, we demonstrate how modern language models can assist in data collection and annotation.

\paragraph{The contributions of this paper are as follows:}

\begin{itemize}
\item Presenting a real-world dataset annotated by clinical experts, focusing on the language and speech deficiencies of individuals with bipolar disorder and schizophrenia.
\item Introducing a model that assists clinicians in maintaining dialogue with recruited participants for data collection purposes.
\item Developing another model that annotates real participant data based on domain-specific variables.
\item Demonstrating that our models achieve low error rates and higher accuracy compared to commercial language models like GPT-4.
\end{itemize}

\section{Data Collection and Labeling}
\label{data_coll}

We start by using the dataset introduced by \citet{aich-etal-2022-towards} in 2022. The data consists of transcribed texts from interviews with 644 participants. In the initial dataset, the authors recruited participants from three categories: participants with schizophrenia, participants with bipolar disorder, and healthy control groups. The diagnoses for subjects are all based on the DSM-V. To build the dataset, the participants were in two simulated clinical tasks with expert clinicians. For task descriptions, please refer to appendix \ref{appendix:Task_Desc}.

We present a clinical annotation task to expand the dataset. 

\subsection{Clinical Annotation of Data}
\label{sec:clinical_scores}

We collect clinical scores for our SSPA data. The SSPA instrument variables \citep{SSPA_instruments} are defined below. Annotators adhering to these definitions were found to have near-perfect agreement \{$\kappa \geq 0.85$\} when labeling the presence of these variables \citep{PATTERSON2001351}:

\begin{itemize}
    \item \textbf{Interest/Disinterest:} Subjects with a relevant mental health condition show low engagement in SSPA tasks since functions of the brain are impaired.
    \item \textbf{Fluency:} Subjects with higher fluency use fewer filler words such as \textit{umm}, \textit{you know}, or \textit{sooo}, and/or fewer long pauses during SSPA tasks.
    \item \textbf{Clarity:} Subjects with greater communication clarity exhibit stronger coherence in speech, both in how things were said and what was said. In lay terms, this variable describes how well subjects are able to get their point across.
    \item \textbf{Focus:} Subjects with greater focus are able to more solely concentrate on the task that has been given to them without veering from their course. This variable also describes the subject's ability to maintain attention on the interviewer and the current and overall task objectives.
    \item \textbf{Social Appropriateness:} Subjects with greater social appropriateness scores fare better socially with respect to the scene. They react more appropriately to interview cues and are able to maintain increased composure during tasks.
\end{itemize}

These five scores for the SSPA are based on the interactions of participants with the clinicians. Each of these scores is annotated for a subject in each scene. The scores are then averaged across the scene for the subject. A subject's total SSPA score is the average of their two scene scores. Scoring is performed manually by experts, achieving a high inter-class coefficient. As shown in prior work \citep{PATTERSON2001351}, subjects' SSPA scores are significantly correlated with the presence or absence of schizophrenia/schizoaffective disorder ($p < 0.01$).\footnote{Results are from a t-test taken comparing SSPA scores for schizophrenia and control group patients.}

For annotation and collection, there were two expert annotators. These were practicing clinicians and researchers in psychiatry. Each annotator reviews the entire transcript and labels all five scores. Gold standard labels are adjudicated by discussion among clinical experts. The SSPA is a well-established standardized test with the scoring metrics clearly defined. Cohen's Kappa $\kappa$ for all clinical scores was $\kappa \geq 0.85$. For our work, we consider the final adjudicated gold standard labels.

\section{Methods - Interview Sequence Generation}
\label{methods}

\subsection{Context-Aware Interviewer}
\label{sec:seq2seq}

Our first specialized objective was to design a proof-of-concept context-aware interviewer to facilitate SSPA sessions. Currently the SSPA is administered by human clinicians with heavy case-loads. The US mental healthcare system is already heavily over burdened with a very low number of clinicians to a high number of patients \citep{barriers_to_mh_care}, potentially leading to mistakes and reduced efficiency. Having a trustworthy and viable agent can help alleviate some of this. To administer the SSPA in a language modeling setting understanding of context is important. Each response from an interviewer depends not only on the previous turn, but the entire dialogue history to that point, i.e. the entire context window of that string. As described previously, our SSPA data is represented as two sets of dialogues (lists of $n$ utterances), one of which belongs to the patient $P$ and the other of which belongs to the interviewer $I$: $P = \{P_0, P_1, ...,  P_n\}$ and $I = \{I_0, I_1, ..., I_n\}$.  Both are stored with associated timestamps indicating when utterances begin.

In a real world setting, it is expected that an interviewer has facilitated many interviews before, across people with bipolar disorder and schizophrenia as well as people with neither condition. It is also expected that in each complete dialogue turn $\{P_i, I_i\}$, the Interviewer response $I_i$ is not only a response to the dialogue $P_i$ but to the set of dialogues $\{P_0, I_0, P_1, I_1, ..., I_{i-1}, P_i\}$. The intuition is thus that the interviewer is responding not only to what was just uttered by the patient, but in a way that is suitable with the entire conversation so far, including all patient and interviewer utterances up to the most recent patient utterance.

\subsection{Task Setup for Interview SFT Experiment}

In this section we describe our setup for the supervised fine-tuning (SFT) experiment. We model this task as a sequence to sequence problem. Our model is trained to generate an appropriate sequence of dialogue in response to dialogue sequences it has seen in such a way that it is aligned with that generated by a real-world interviewer.  We train on 75\% of all BD, HC, and SZ dialogues across both scenes. The input and outputs for the encoder-decoder forward pass are: 

\[
    I\rightarrow Out= 
\begin{cases}
    P_0 \rightarrow I_0,& \text{if } n=0\\
    {P_0, I_0, ..., I_{i-1}, P_i} \rightarrow I_i, & n = i
\end{cases}
\]

The equation above again emphasizes that to create input-output pairs we consider the dialogue history in addition to the most recent utterance. If we are at index 0 of a conversation, the interviewer's response is based directly on the the patient's utterance, but otherwise the interviewer response is based on the entire dialogue history between the patient and interviewer, until the \textit{i-1}th interviewer utterance and the patient utterance $P_i$. 

\begin{figure*}[h]
    \centering
    \fbox{\includegraphics[scale=0.6]{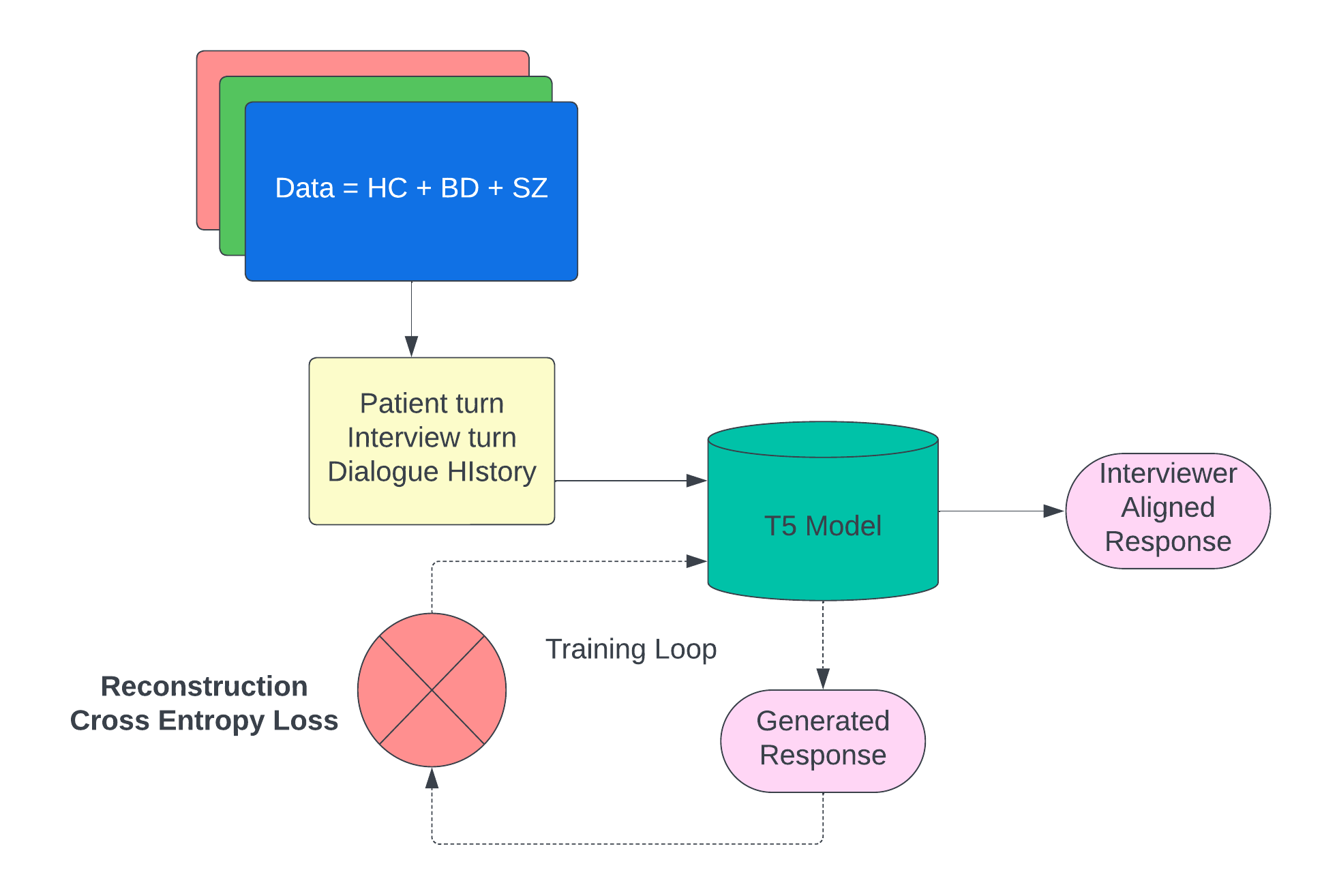}}
    \caption{Interview model turns and dialogue history to calculate reconstruction loss and generate well aligned sequences towards the SSPA}
    \label{fig:interview_model}
\end{figure*}

A schematic diagram for the SSPA language modeling process is shown in Figure \ref{fig:interview_model}. The model is fine-tuned until the loss drops from 1.64 initially to 0.1 after 15000 checkpoints and then results are calculated. To initialize training we provide the following source prefix: 

\begin{quote}
\texttt{You are an intelligent interviewer see the 
examples provided and learn to interview a new patient}
\end{quote}

We selected this prefix after experimenting with simpler versions (e.g., \textit{"Interview a patient"} and \textit{"Talk to a patient based on examples"}) and finding that the more complex final prefix was necessary to produce results well-aligned with reference interviews. This process does not involve in-context learning (ICL), prompt engineering, or tuning; it is a manually constructed prefix. Current literature suggests that better prefix descriptors, followed by improved training, yield superior results \citep{xue-etal-2022-byt5}. Standard hyperparameters were maintained at default values, and training was conducted on four 16GB T4 GPUs.

The fine-tuned model was tested individually on all scenes and classes to simulate real-world interview conditions, where the interviewer focuses on a single scene and person, independent of prior interview training. To evaluate the quality of generated output, we computed syntactic similarity using ROUGE \citep{lin-2004-rouge} scores, semantic similarity using cosine similarity, and alignment with human dialogue using BERTScore \citep{zhang2020bertscore}.

\subsection{Results: Generated Interview Quality}

In Table \ref{tab: sft_interview} we present the results of the interview SFT experiment. To compute semantic similarity scores, we first encoded both the generated utterance and the corresponding gold expected utterance as \texttt{deberta} embeddings owing to the model's increased ability to align with human speech \citep{zhang2020bertscore,he2021debertav3}, and then calculated the cosine similarity between those embeddings. To compute syntactic similarity, we use ROUGE-1 to find overall unigram overlap and ROUGE-L to find the longest common sub-sequence overlap.  We report precision, recall, and F1 scores for these two metrics using the ROUGE-SCORE package from the Python library.\footnote{\url{https://pypi.org/project/rouge-score/}} We use the BERTScore \citep{zhang2020bertscore} metric directly, using a deberta model to vectorize the inputs to the metric generator \citep{zhang2020bertscore},\footnote{BERTScore needs users to specify which model to use to calculate metrics between two given strings.  We use deberta for the same reasons cited earlier; i.e., studies have found it to generate text that more closely matches human speech.} and report the precision, recall, and F1-score.  According to the authors of the original paper, this model (deberta) offers the best understanding of the closeness of generated text to human intent. 
For all semantic metrics we use deberta as our choice of model since it has been consistently shown to outperform other encoder based popular choices like BERT or RoBERTa \citep{he2021debertav3}. 

\begin{table*}[t]
\centering
\small
\begin{tabular}{lcccccccccc}
\toprule
Class $\times$ Scene& Semantic& \multicolumn{6}{c}{Syntactic Similarity}                                                                                                                                                 & \multicolumn{3}{c}{Human Alignment}                                    \\ \cmidrule(lr){2-2} \cmidrule(lr){3-8} \cmidrule(lr){9-11}
        &                                    Cosine& \multicolumn{3}{c}{ROUGE-1}                                                           & \multicolumn{3}{c}{ROUGE-L}                                                           & \multicolumn{3}{c}{BERTScore}\\ \cmidrule(lr){2-2} \cmidrule(lr){3-5} \cmidrule(lr){6-8} \cmidrule(lr){9-11}
          &          & {P} & {R} & {F1} & {P} & {R} & {F1} & {P}        & {R}     & F1\\ \midrule
BD Scene\_1                     &                                    0.652& \multicolumn{1}{|l|}{0.381}          & \multicolumn{1}{|l|}{0.380}       & \multicolumn{1}{|l|}{0.360}   & \multicolumn{1}{|l|}{0.363}          & \multicolumn{1}{|l|}{0.370}       & \multicolumn{1}{|l|}{0.340}   & \multicolumn{1}{|l|}{0.66}          & \multicolumn{1}{|l|}{0.66}       &    0.66\\ \midrule
BD Scene\_2                     &                                    0.623& \multicolumn{1}{|l|}{0.361}          & \multicolumn{1}{|l|}{0.346}       & \multicolumn{1}{|l|}{0.334}   & \multicolumn{1}{|l|}{0.344}          & \multicolumn{1}{|l|}{0.336}       & \multicolumn{1}{|l|}{0.317}   & \multicolumn{1}{|l|}{0.61}          & \multicolumn{1}{|l|}{0.61}       &    0.61\\ \midrule
SZ Scene\_1                     &                                    0.634& \multicolumn{1}{|l|}{0.331}          & \multicolumn{1}{|l|}{0.316}       & \multicolumn{1}{|l|}{0.301}   & \multicolumn{1}{|l|}{0.328}          & \multicolumn{1}{|l|}{0.314}       & \multicolumn{1}{|l|}{0.300}   & \multicolumn{1}{|l|}{0.63}          & \multicolumn{1}{|l|}{0.64}       &    0.63\\ \midrule
SZ Scene\_2                     &                                    0.613& \multicolumn{1}{|l|}{0.371}          & \multicolumn{1}{|l|}{0.362}       & \multicolumn{1}{|l|}{0.346}   & \multicolumn{1}{|l|}{0.360}          & \multicolumn{1}{|l|}{0.352}       & \multicolumn{1}{|l|}{0.340}   & \multicolumn{1}{|l|}{0.62}          & \multicolumn{1}{|l|}{0.63}       &    0.61\\ \midrule
HC Scene\_1                     &                                    0.670& \multicolumn{1}{|l|}{0.390}          & \multicolumn{1}{|l|}{0.390}       & \multicolumn{1}{|l|}{0.360}   & \multicolumn{1}{|l|}{0.380}          & \multicolumn{1}{|l|}{0.390}       & \multicolumn{1}{|l|}{0.360}   & \multicolumn{1}{|l|}{0.67}          & \multicolumn{1}{|l|}{0.68}       &    0.67\\ \midrule
HC Scene\_2                     &                                    0.643& \multicolumn{1}{|l|}{0.402}          & \multicolumn{1}{|l|}{0.392}       & \multicolumn{1}{|l|}{0.380}   & \multicolumn{1}{|l|}{0.390}          & \multicolumn{1}{|l|}{0.380}       & \multicolumn{1}{|l|}{0.370}   & \multicolumn{1}{|l|}{0.64}          & \multicolumn{1}{|l|}{0.64}       &    0.63\\ \bottomrule
\end{tabular}
\caption{Interview SFT Results.  \textit{P}=precision, and \textit{R}=recall.}
\label{tab: sft_interview}
\end{table*}


BERTScores, designed to capture intent and semantic similarity, are almost double the corresponding ROUGE scores for the same scenes. Recent studies have shown \citep{zhang2020bertscore,hanna-bojar-2021-fine} that BERTScore has two important properties. Firstly, it correlates with other summarization and similarity metrics (e.g., cosine similarity or BLEU score). Secondly, when a task becomes harder such as in our case, BERTScore accuracy peaks around 80\% \citep{hanna-bojar-2021-fine}. Considering that our BERTScores for our task are close to 70\% we can conclude that our model works at a high performance level. A better cosine similarity represents closeness in the embedding space of the vectors, whereas a good BERTScore tells us that the outputs are aligned with the reference sample. 

However, even with a well-performing model, our ROUGE score is quite low. Some of this may be attributed to hallucinatory effects. For example, we observe that in one case while the interviewer in the original script says, e.g., "My name is \textsc{Interviewer}," our model generates, e.g., "My name is \textsc{Name}"---that is, a hallucinated name that was never previously mentioned in the dialogue.  Thus, although this is structurally aligned with the reference, it differs in a key way that is best captured by ROUGE. 

Another reason why our model exhibited lower syntactic than semantic performance may lie in disfluency.  In our reference dialogues the interviewers often pause using filler words like \textit{uh}, \textit{uhh}, \textit{okay}, or \textit{mmhmm} to give the patients more time to speak. While our model thematically aligns decently well with these statements, its exact filler word matches are quite low. For example, we observe that the model also pauses but uses different filler words, or longer sequences of filler words, negatively affecting our ROUGE metric. However, throughout our observations, we can see that the model seems to understand the SSPA expectation of the interviewer role, even though we do not specify this in our SFT setup explicitly. 

We qualitatively observe that the model appears capable at staying on-topic for the scene-specific task (e.g., generating content like \textit{"Of course I will try to send someone over the fix the leak."}). It is interesting to observe that the model can discern the underlying task over long periods of training. Even without telling the model explicitly what the SSPA task involves, we can see that the model understands that a leaky pipe is at its core. This may suggest that LLMs are well-suited for tasks with better data and longer training \citep{min2022metaicl, brown2020language}. While our alignment-based scores are not exceptionally high, this is still a strong starting benchmark for a nuanced task \citep{hanna-bojar-2021-fine}. The model captures close to 70 points of alignment with the intent of the actual interviewer. In the next phase we use this to generate annotator scores using another model to further progress the autonomous pipeline.

\section{Methods - Annotation Generation}
\label{sec:score_pred}

We also frame our SSPA score prediction task as a sequence to sequence task. Rather than simply predicting a sequence of scores, we also predict the label for which the score is being generated. In Section \S{}\ref{sec:clinical_scores} we discussed what the five clinical variables are and how the scores are collected. In this score prediction task, the model learns to predict the score (SSPA clinical value) and the corresponding label. Therefore, the model predicts a sequence $Interest=XX, Fluency=YY$ rather than a simple distribution $4,5,3..$. This increases complexity, but helps us evaluate and walk towards a more explainable model. 
For this setting we use the interview dialogues from our source dataset and a \texttt{t5-base} LLM.  We use the following prompt to generate scores:

\begin{quote}
\texttt{You are an intelligent annotator see the 
examples provided and generate scores for each variable}
\end{quote}

The source prefix selection and other model parameters are kept the same as in the interview generation task described earlier in this paper. The model trains for 10000 checkpoints and the validation loss goes from 0.8 to 0.02 in our best performing model checkpoint. We calculate this reconstruction loss between the variable labels and scores that are annotated by our clinicians and the ones that are generated by our model. A standard cross-entropy loss function is used to find the loss. The model is trained on 75\% of the data and validated on 5\% of the data, with the remainder held out for testing.

\subsection{Results: SSPA Score Prediction}
\label{sec:result_score}

The results of the SSPA score prediction model are presented in Table \ref{tab:control_score}. The values represent the root mean squared error (RMSE) between the original annotated labels \( Y = \{S_1, S_2, \ldots, S_n\} \) and the predicted labels \( Y' = \{S'_1, S'_2, \ldots, S'_n\} \), where \( Y \in \{ \text{Interest}, \text{Fluency}, \text{Clarity}, \text{Focus}, \text{Social} \} \). The results indicate generally low error, with improved predictive performance in Scene 2 compared to Scene 1.

The model exhibits superior performance for the variables \textit{Social} and \textit{Focus}, which is anticipated as the SSPA predominantly evaluates social skills, and \textit{Social} encapsulates social appropriateness. The variables \textit{Focus}, \textit{Clarity}, and \textit{Fluency} are linguistically dependent, with the model performing best in this order, and the least effective for \textit{Interest}. The higher RMSE for \textit{Interest} can be attributed to its reliance on non-verbal cues such as body language, which are absent from our transcripts.

Overall, this standalone model demonstrates effective prediction capabilities. In the subsequent section, we illustrate the adaptation of our previous model from \S\ref{sec:seq2seq} into a chained pipeline, enabling SSPA interview transcripts to be scored with minimal RMSE differences compared to the standalone model.

\begin{table*}[t]
\centering 
\begin{tabular}{lllllll}
\toprule
Class and Scene & \multicolumn{5}{l}{RMSE} &                        \\ \midrule
                & \textbf{Interest} & \textbf{Fluency} & \textbf{Clarity} & \textbf{Focus} & \textbf{Social}  &\textbf{Avg.~RMSE/Case}\\
                \midrule
BD Scene\_1    & 1.36     & 1.10    & 1.04    & 0.97  & 1.06    & 1.10\\
BD Scene\_2    & 1.09     & 1.11    & 1.14    & 1.15  & 1.12    & 1.12\\
SZ Scene\_1    & 1.27     & 1.27    & 1.28    & 1.19  & 1.30    & 1.26\\
SZ Scene\_2    & 1.22     & 1.10    & 1.13    & 1.10  & 1.07    & 1.12\\
HC Scene\_1    & 1.28     & 1.36    & 1.35    & 1.33  & 1.33    & 1.33\\
HC Scene\_2    & 0.84     & 0.78    & 0.68    & 0.84  & 0.68    & 0.76\\
 Avg.~RMSE/Var& 1.17 & 1.12 & 1.10 & 1.09 & 1.09 & N/A
 \\ \bottomrule
\end{tabular}
\caption{RMSE scores for standalone score prediction model, using original dataset. Avg-RMSE/Case represents the mean RMSE across a diagnostic group and scene. Avg-RMSE/Var represents the mean RMSE for that SSPA variable of the column.}
\label{tab:control_score}
\end{table*}

\section{Chained Model}

\begin{table*}[htbp]
\centering 
\begin{tabular}{lllllll}
\toprule
Class and Scene & \multicolumn{5}{l}{RMSE}                        \\ \midrule
                & \textbf{Interest} & \textbf{Fluency} & \textbf{Clarity} & \textbf{Focus} & \textbf{Social} & \textbf{Mean/Case}\\
                \midrule
BD Scene\_1    & 1.28& 1.12& 1.07& 0.97& 1.06& 1.10\\
BD Scene\_2    & 1.39& 1.11& 1.14& 1.18& 1.10& 1.18\\
SZ Scene\_1    & 1.37& 1.33& 1.27& 1.20& 1.30& 1.29\\
SZ Scene\_2    & 1.33& 1.13& 1.12& 1.15& 1.10& 1.16\\
HC Scene\_1    & 1.33& 1.37& 1.27& 1.30& 1.28&1.31\\
HC Scene\_2    & 0.83& 0.78& 0.75& 0.92& 0.75&0.80\\ 
Avg.~RMSE/Var   & 1.25& 1.14&1.10&1.12&1.09& N/A\\
\bottomrule
\end{tabular}
\caption{RMSE scores for the chained score prediction model.  Interview sequences come from the generative model described in \S{}\ref{sec:seq2seq}. Mean/Case represents the mean RMSE across a diagnostic group and scene. Avg-RMSE/Var represents the mean RMSE for that SSPA variable of the column.}
\label{tab:continue_score}
\end{table*}

So far in this paper we have created two standalone models: one in \S{}\ref{sec:seq2seq} that can learn from interviewers to appropriately interact with patients to facilitate the SSPA task, and the other in \S{}\ref{sec:score_pred} that reads patient-interviewer transcripts and generates a sequence of SSPA scores for a patient. In this section we experiment with combining them. The primary motivation for this lies in anticipated real-world need, moving towards a seamless support tool for busy clinicians who may otherwise need to administer and score the SSPA manually. We create a chained model that (1) converses with the patient, and (2) predicts SSPA scores from the encounter. 

\begin{figure*}[t]
    \centering
    \fbox{\includegraphics[scale=0.5]{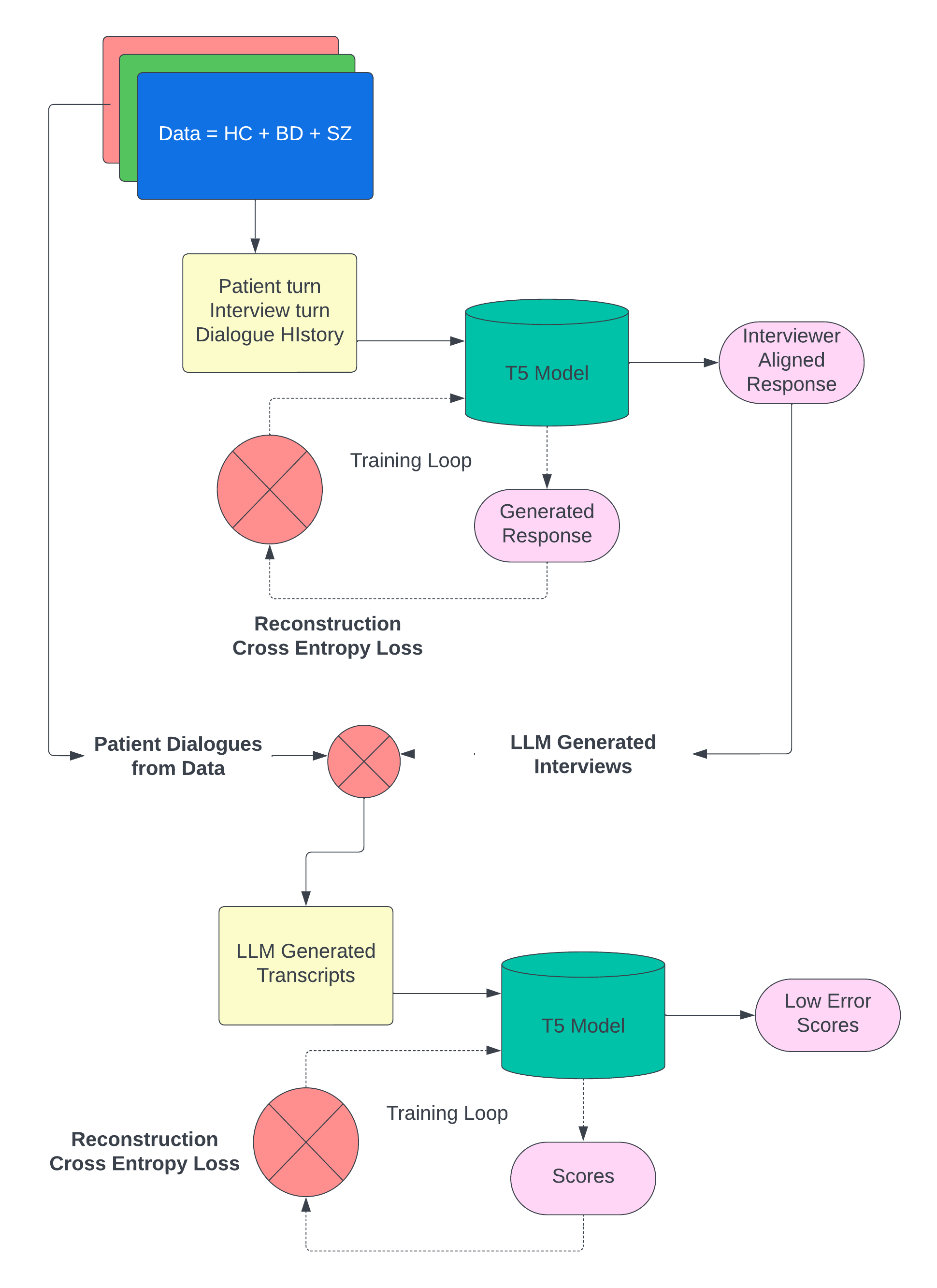}}
    \caption{Chained Model Setup. Two standalone t5 models are chained by output and input. The Interview generator model works with patient dialogues to create LLM generated transcripts. This is fed into the score prediction model which outputs low error scores for the SSPA using a cross-entropy loss function.Picture resized for space limitations. Please zoom-in while reading review version.}
    \label{fig:chained_model}
\end{figure*}

\begin{table*}[htbp]
    \centering
    \begin{tabular}{cccccc}
    \toprule
      \textbf{BD Scene 1} & \textbf{BD Scene 2} & \textbf{SZ Scene 1} & \textbf{SZ Scene 2} & \textbf{HC Scene 1} & \textbf{HC Scene 2}\\
      \midrule
        0.0 & 0.06 & 0.03 & 0.04 & 0.2 & 0.04\\
        \bottomrule
    \end{tabular}
    \caption{Difference of mean errors per case and scene.}
    \label{tab:rmse_difference_case}
\end{table*}

\begin{table}[]
    \centering
    \begin{tabular}{ccccc}
    \toprule
      \textbf{Interest} & \textbf{Fluency} &  \textbf{Clarity} & \textbf{Focus} & \textbf{Social}\\
      \midrule
        0.8 & 0.02 & 0.0 & 0.03 & 0.0 \\
        \bottomrule
    \end{tabular}
    \caption{Difference of mean errors per variable.}
    \label{tab:rmse_difference_var}
\end{table}

We predict scores for dialogues that our model generated in \S{}\ref{sec:seq2seq}. The input consists of the entire dialogue between the patient $P$ and generated interviewer dialogues $I_{gen}$, forming the sequence $\{P_0, I_0, P_1, I_1, ... P_n, I_n\}$, where an interviewer dialogue $I_k \in \{I_{gen}\}$ acts as input and the model returns a sequence of five integer-valued scores, $\{S_1,S_2,..,S_5\}$, that quantify the SSPA variables defined in \S{}\ref{sec:clinical_scores} (Interest, Fluency, Clarity, Focus, and Social).

\subsection{Results: Chained Model}

We present the results of the experiment in Table \ref{tab:continue_score}. The scores reported are the RMSE between the expected SSPA scores and the generated SSPA scores predicted for LLM-facilitated SSPA transcripts. Our first observation is the acute closeness to the stand-alone model scores (recall Table \ref{tab:control_score}). This shows that even when LLM-based assistants are adapted in a chained end-to-end fashion, the results are similar to those observed using standalone models.

When we compare the difference between errors for Tables \ref{tab:control_score} and \ref{tab:continue_score} we can see that the differences are quite low at both a variable level and a \textit{class X scene} level. We can see in Tables \ref{tab:rmse_difference_var} and \ref{tab:rmse_difference_case} that on a per scene or per variable basis the differences are quite low with no significant difference \footnote{A t-test between the RMSE scores per case (scene and class) and per variable shows the differences between score distributions for the standalone and chained models are not statistically significant (p < 0.05).}

\section{Baseline Comparison and Discussion}
\label{sec:Baseline_Discuss}

Below we provide a baseline comparison between GPT-4, GPT-4o, and our method in replicating annotation tasks as detailed in \S{}\ref{sec:clinical_scores}. The results, presented in Table \ref{tab:baseline}, illustrate the mean errors per class and scene, with statistical significance validated using the Wilcoxon signed-rank test. We find that for domain-specific clinical variables, GPT and GPT-4o show high errors in generating appropriate scores and show a high degree of error. We can see that when compared to our model we get statistically significant low errors showing the need for appropriate fine-tuning. 

\begin{table}[!ht]
\resizebox{250 pt}{60 pt}{\begin{tabular}{@{}llllll@{}}
\toprule
Scene/Class & GPT-4 Error & GPT-4o Error & Our Error & p 4 & p - 4o \\ \midrule
HC - Sc - 1 & 1.60 & 1.57 & 0.2 & 0.03 & 0.03 \\
HC - Sc -2 & 1.70 & 1.66 & 0.04 & 0.03 & 0.03 \\
SZ - Sc -1 & 1.44 & 1.50 & 0.03 & 0.03 & 0.03 \\
SZ - Sc - 2 & 1.64 & 1.53 & 0.04 & 0.03 & 0.03 \\
BD - Sc - 1 & 1.51 & 1.49 & 0.00 & 0.03 & 0.03 \\
BD - Sc - 2 & 1.45 & 1.53 & 0.05 & 0.03 & 0.03 \\ \bottomrule
\end{tabular}}
\caption{Baseline Comparison with GPT4 and GPT4o. We can see that the p values comparing our error with GPT errors show a significant difference. }
\label{tab:baseline}
\end{table}

Our experiments reveal two significant trends in our interview replication model: an intrinsic comprehension of tasks and the generation of unrelated information. Even without explicit task instructions, a well-constructed fine-tuning loop allows a smaller model to intuitively understand tasks, evidenced by the model's ability to identify tasks from indirect references. Despite the tendency of the model to hallucinate information such as names and dates, which typically impedes performance on tasks necessitating factual precision, our findings indicate that these hallucinations do not compromise task completion. For our annotation task, we maintained a sequence-to-sequence setup for predicting scores and variable labels, observing low error rates and consistent performance across both stand-alone and chained model setups. 
\section{Conclusion}
\label{sec:conclusion}

In this paper, we present a collection of clinically enriched data, which is subsequently annotated by experts across five major categories, achieving high agreement. We demonstrate how a modern language model architecture can aid in the automation of data collection and annotation. Unlike the original authors of the dataset, we do not conduct classification experiments. Furthermore, we refrain from making any claims that language models should be used for medical diagnoses or clinical conclusions regarding an individual's mental health or medical status.

\paragraph{The main contributions of this paper are as follows:}
\begin{itemize}
\item Demonstrating the practical application of generative AI to address significant clinically-enriched healthcare challenges.
\item Illustrating how fine-tuned Seq2Seq models can be employed for data collection through standardized tasks and clinical annotation with minimal error.
\item Highlighting the limitations of contemporary large language models, such as GPT-4 and GPT-4o, in capturing the nuances of domain-specific variables, and showing how these gaps can be addressed with smaller models and targeted fine-tuning.
\end{itemize}

Our findings indicate that language models can assist clinicians in scaling data collection and labeling with a high degree of trust, as evidenced by low error rates and high similarity scores. We anticipate that the clinical community will find our models ready for implementation and our methods both translatable and scalable for their specific tasks.

\section{Limitations}
\label{sec:limit}

In this study, we worked with 644 participants, a relatively small sample size. In their original work, \citet{PATTERSON2001351} identify eight variables of the SSPA, but we selected only five for this task. The excluded variables were either independent of speech (e.g., personal grooming) or lacked expert raters due to their independence from healthcare (e.g., negotiation ability). We utilized the T5 model due to hardware limitations. Despite its smaller size, the T5 model demonstrated that even with limited computational resources, it could achieve relatively low error rates. This suggests that a larger model with more computational power could potentially reduce errors further. Additionally, our study is confined to transcripts of audio recordings from the original data, without incorporating multimodal aspects.

\section{Ethical Concerns}
\label{sec:ethics}

This paper aims to demonstrate how modern language models can be deployed in clinical settings to collect and label data responsibly. We exclusively use labels that are well-established in clinical contexts. Importantly, this paper does not advocate for or implement the use of language models as diagnostic tools for mental health. We illustrate that markers of speech relevant to psychiatric healthcare can be predicted using language models. However, predicting variables like Interest or Focus should not be used or interpreted for unrelated tasks, such as advertising, targeted marketing, or any clinical purposes without appropriate expertise.

All data-related activities, including labeling, annotation, and sharing, were conducted with the approval of four independent academic Institutional Review Boards (IRBs). Participants in the original study provided informed consent. We adhere to all ethical codes established by the ACM and ACL. This paper involves numerous clinical experts in the labeling, adjudication, and language modeling processes, ensuring proper guidance and assistance. Using these models or concepts from this paper for non-clinical purposes or without expert guidance in clinical contexts is strictly prohibited.

\section*{Acknowledgements}

This research was supported in part by the Intramural Research Program of the NIH, National Institute on Drug Abuse (NIDA). Research reported in this publication was also partially supported by the National Institute of Mental Health of the National Institutes of Health (NIMH) under award number R01MH116902.

\bibliography{custom}

\appendix

\section{Task Description and Purpose}
\label{appendix:Task_Desc}

In this appendix we briefly describe the Social Skills Performance Assessment. We will talk about the purpose of the task, the task itself, and what we can gain from this task. 

\paragraph{Task Motivation}

The Social Skills Performance Assessment, \textit{abbrev. SSPA} is a mental-health instrument which serves as an indicator of social skill. The motivations, some of which we discussed in the introduction, is that people with psychiatric illnesses are more likely to show less cohesion and more disorganization in their speeches, as opposed to healthy control subjects. The SSPA task standardizes the way speech is measured for subjects with, and without psychiatric illnesses by having the subjects take on two tasks with expert clinicians. 

For both tasks, the participants speak with a trained clinician. Their video and audio are recorded. Then transcribed. The labels mentioned in this paper were the clinicians rating the participants performance on the tasks to the two tasks. 

\paragraph{Task Description}

There are two tasks to the SSPA. The first task is the neutral or friendly task, and the second task is the confrontational task. 

\paragraph{The Friendly Task} consists of the participant simulating a conversation as if they moved to a new neighborhood. They are asked to introduce themselves to the new neighbor. We observe people without psychiatric illnesses to briefly talk about 2-3 topics and stay consistent. People with illnesses tend to sway between 13-15 topics and are unable to concisely present thoughts. 

\paragraph{The Confrontational Task} consists of the participant complaining to their landlord after a leaky pipe has not been fixed for months. We observe that healthy controls are able to quickly articulate and talk only about the problem at hand. We observe that BD and SZ often talk about multiple different things and then talk about the problem given to them. 

\paragraph{Task Outcomes}

Annotating clinical variables is a different task than classification. While these variables are not classifiers of psychiatric illnesses. They are important features. These variables give clinicians and scientists much needed quantification in the field of life-long psychiatric illnesses. Therefore, it is imperative to bring modern technology to the equation and slowly make care and data collection accessible and efficient. 

\end{document}